\documentclass[lettersize,journal]{IEEEtran}

\usepackage[utf8]{inputenc}
\usepackage{amsmath}
\usepackage{mathtools}
\usepackage{amssymb,amsfonts}
\usepackage{array}
\usepackage{graphicx}
\graphicspath{{./figures/}{../figures/}}
\usepackage{booktabs}
\usepackage{cite}
\usepackage{algorithm}
\usepackage{algorithmic}
\usepackage{multirow}
\usepackage{bm}
\usepackage{xcolor}
\usepackage[hidelinks]{hyperref}
\usepackage{makecell}

\DeclareMathOperator{\softmax}{softmax}

\DeclareMathOperator{\GELU}{GELU}
\DeclareMathOperator{\LN}{LayerNorm}
\DeclareMathOperator{\MHA}{MHA}
\DeclareMathOperator{\FFN}{FFN}
\DeclareMathOperator{\ReLU}{ReLU}

\newcommand{\etal}{\textit{et al.}}
\newcommand{\ie}{\textit{i.e.}}

\begin{document}
	
	\title{Text-Anchored Modality Experts with Uncertainty-Guided Routing for Multimodal Sentiment Analysis}
	
	\author{
		Xiaohua~Huang,
		Bingrui~Lei,
		Xiuyan~Shao,
		Ling~Zhou,
		Yuan~Zong,
		Wenming~Zheng,
		and Guoying Zhao
		
		\thanks{Xiaohua Huang and Bingrui Lei are with the School  of Communication and Information Engineering, Nanjing Institute of Technology, Nanjing 210096, China.}
		
		\thanks{Xiuyan Shao is with the School of Economics and Management, Southeast University, Nanjing 211196, China.}
		
		\thanks{Ling Zhou is with the School of Computer Science and Engineering, Macau University of Science and Technology, Macau SAR, 999078, China.}
		
		\thanks{Yuan Zong and Wenming Zheng are with the Key Laboratory of Child Development and Learning Science (Southeast University), Ministry of Education, Nanjing 210096, China, and also with the School of Biological Science and Medical Engineering, Southeast University, Nanjing 210096, China.}
		
		\thanks{Guoying Zhao is with the Center for Machine Vision and Signal Analysis, University of Oulu, 90570 Oulu, Finland.}
		
	}
	
	\maketitle
	
	\begin{abstract}
		
		Multimodal sentiment analysis (MSA) aims to understand user-generated videos by integrating multiple modalities. Existing methods often treat modalities as symmetric, overlooking the fact that language provides a dominant semantic prior while nonverbal signals are complementary and noise-prone. To address this, we propose a Text-Anchored Modality Experts with Uncertainty-Guided Routing, termed TAME, that reformulates MSA as an uncertainty-guided refinement paradigm, where the language prior is calibrated and enhanced by selectively trusted nonverbal evidence. Specifically, TAME adopts an asymmetric Mixture-of-Modality-Experts framework, in which textual features act as a fixed and always-activated shared expert, while audio-textual and video-textual cross-modal branches serve as dynamically routed auxiliary experts. Furthermore, we introduce lightweight Prophet Heads to quantify the sample-level predictive uncertainty of each individual modality, and this uncertainty metric is further incorporated into the modality router to realize reliability-aware expert aggregation and fusion.  Experiments on CMU-MOSI, CMU-MOSEI, and CH-SIMS datasets demonstrate that TAME achieves state-of-the-art performance on the majority of metrics while reducing parameter count by up to 78.7\% and FLOPs by up to 41.2\% over the mixture-of-expert baseline.

	\end{abstract}
	
	
	\section{Introduction}
	\label{sec:intro}
	
	Multimodal sentiment analysis (MSA) aims to predict the sentiment expressed in an utterance by jointly leveraging language, acoustic, and visual signals. A dominant paradigm treats the three modalities as symmetric information sources and seeks to align or fuse them as thoroughly as possible, for example via all-pair cross-modal transformers~\cite{tsai2019multimodal}, modality-invariant decomposition~\cite{hazarika2020misa}, or mutual-information maximization~\cite{han2021improving}. Although this symmetric fusion philosophy has driven steady progress, it overlooks a fundamental asymmetry that is well documented in the literature: on text-centric sentiment benchmarks such as CMU-MOSI, language generally provides the strongest and most stable semantic prior, while audio and video contribute information that is complementary but noisier and more sample-dependent~\cite{yu2021learning,das2023multimodal}.
	
	This observation motivates a different question. Rather than asking \emph{“how can we align all modalities more completely?”}, we ask: \emph{when should a model trust nonverbal evidence to revise the initial judgment that language alone provides?} Answering this question at the sample level requires two capabilities that symmetric fusion methods lack: an architectural bias that treats text as a prior rather than an equal participant, and an explicit mechanism for estimating the reliability of each nonverbal evidence source before admitting it.
	
	We address both requirements in a unified framework called \underline{T}ext-\underline{A}nchored \underline{M}odality \underline{E}xperts with Uncertainty-Guided Routing (TAME). Motivated by the shared-plus-routed expert paradigm from recent LLMs~\cite{dai2024deepseekmoe}, TAME treats the text branch as an always-on shared expert that anchors the semantic prior, while the audio–text and video–text branches serve as routed experts whose contributions are dynamically gated by per-sample uncertainty estimates from lightweight Prophet Heads. The outcome is an “evaluate-then-trust” decision process, in which the model first assesses how reliable each nonverbal signal is, then decides how much it should be allowed to revise the language prior.	In summary, our main contributions are as follows:
	\begin{itemize}
		\item TAME designates the text branch as an always-on shared expert and the audio–text and video–text branches as sparsely activated routed experts. This asymmetric shared-plus-routed expert architecture explicitly reflects the empirically observed text dominance in sentiment benchmarks.
		
		\item To ground the interpretation of nonverbal signals in reliable linguistic context, we introduce a text-anchored cross-modal attention mechanism that restrict cross-attention to audio-to-text and video-to-text pathways, deliberately excluding direct audio–video interactions. This design acts as a semantic filter, preventing spurious nonverbal correlations from bypassing the text anchor and distorting routing decisions.
		
		\item We propose a pyramid attention pooling module to capture sentiment cues at different temporal granularities, together with a heteroscedastic uncertainty estimation strategy that independently quantifies each modality’s aleatoric uncertainty via a Prophet Head. The estimated log-variances are directly injected as a prior bias into the router logits, enabling sample-level, end-to-end reliability-aware fusion without external calibration.
		
		\item Extensive experiments on three benchmarks demonstrate that TAME consistently outperforms or achieves competitive performance among state-of-the-art methods, validating the effectiveness of text-anchored cross-modal attention, multi-scale temporal pooling, and uncertainty-aware routing.
		
	\end{itemize}


	\section{Related Work}
	\label{sec:related}
	
	\subsection{Multimodal Sentiment Analysis}
	
	Multimodal sentiment analysis (MSA) focuses on predicting sentiment by jointly modeling textual, visual, and auditory signals. Early approaches rely on tensor-based fusion strategies. For example, Tensor Fusion Networks (TFN)~\cite{zadeh2017tensor} and Low-rank Multimodal Fusion (LMF)~\cite{liu2018efficient} capture inter-modal interactions through outer products, yet they suffer from limited representation flexibility. With the advancement of Transformer architecture, cross-modal attention has become the mainstream way for MSA. MulT~\cite{tsai2019multimodal} introduces all-pair directional cross-modal attention to align multimodal sequences. MISA~\cite{hazarika2020misa} decomposes features into modality-invariant and modality-specific components to enhance representation robustness. Furthermore, Self-MM~\cite{yu2021learning} exploits self-supervised unimodal learning to preserve modality-specific information, while MMIM~\cite{han2021improving} enhances fusion quality by maximizing mutual information across modalities. Other advances include CRIL~\cite{yang2022disentangled} for contrastive cross-modal alignment, DMD~\cite{li2023decoupled} for modeling decouples multimodal dynamics, and DLF~\cite{wang2025dlf} for exploring adaptive deep late fusion. Despite this progress, most existing methods assume complete and aligned inputs, which is often unrealistic. To address missing modalities and unaligned sequences, Huan~\etal~\cite{huan2024unimf} proposed UniMF, a lightweight Transformer framework that generates missing modalities via a specialized attention mask and fuses information through a learnable multimodal sequence. Furthermore, to mitigate spurious correlations that harm out-of-distribution generalization, Huan~\etal~\cite{huan2025muldef} presented MulDeF, a model-agnostic debiasing framework combining causal intervention and counterfactual reasoning to remove label bias and multimodal biases. 
	
	Nevertheless, these methods still fail to fully exploit the modality-specific information that is also important for MSA. Some recent methods have addressed this issue by projecting the multimodal features into multiple subspaces~\cite{huang2024tmbl} or by using multi-task learning to share knowledge across different sentiment tasks~\cite{zhang2023multitask}. However, both strategies still treat modalities under a largely symmetric structure, either by decomposing a shared feature space or by sharing a unified backbone, and they do not explicitly model the asymmetric roles that different modalities play in text-centric sentiment benchmarks. Moreover, they lack a principled mechanism to estimate the reliability of each modality on a per-sample basis, which is essential for deciding when nonverbal evidence should be trusted.
	
	\subsection{Mixture of Experts in Multimodal Learning}
	
	The Mixture of Experts (MoE) framework~\cite{jacobs1991adaptive} improves model capacity by activating different specialized sub-networks for different input patterns, guided by a learnable gating mechanism. A recent architectural advance is the shared-plus-routed expert design introduced in 
	DeepSeekMoE~\cite{dai2024deepseekmoe}, where a subset of experts is always active,~\ie, shared experts, while others are sparsely activated,~\ie, routed experts. This paradigm has proven effective in large language models~\cite{jiang2024mixtral} for balancing capacity and efficiency.
	
	MoE has also been explored in multimodal learning, for instance in vision–language models for task‑level routing~\cite{lin2026moe}. Within MSA, a key challenge is that modalities are inherently asymmetric: language typically serves as the dominant semantic carrier, while audio and visual cues provide complementary but variable support. Existing MoE-based MSA methods, such as EMOE~\cite{fang2025emoe} and MMoLRE~\cite{zhang2025multimodal}, treats each modality as an independent expert, using a dynamic gating mechanism to adaptively fuse modality‑specific predictions. However, they still assume a symmetric expert organization, treating all modalities as peers and failing to distinguish between generally useful information, which should be always active, and conditionally useful information, which should be selectively activated. This motivates an asymmetric expert design that explicitly anchors on the dominant text modality.
	
	\subsection{Uncertainty in Multimodal Sentiment Analysis}
	
	Generally, uncertainty in deep learning can be categorized into epistemic uncertainty and aleatoric uncertainty~\cite{hullermeier2021aleatoric}. The former represents the uncertainty inherent in the neural network itself, indicating the confidence level in its predictions. The latter refers to the inherent noise present in the data, such as image blurriness, text ambiguity, or missing modalities, which pose significant challenges to the model's predictions. 
	
	Several recent works have introduced uncertainty‑aware learning into MSA. Han~\etal~\cite{han2022trusted} use Dirichlet‑based evidential uncertainty for classification‑level fusion. U‑MoE~\cite{xiao2026multimodal} models both data and cognitive uncertainty via cross‑modal density estimation and uses symmetric KL divergence to quantify modality discrepancies, dynamically adjusting modality weights based on estimated missing rates. EUAR~\cite{gao2024enhanced} proposes an uncertainty‑aware mixture of experts, where enhanced experts capture aleatoric uncertainty and an uncertainty‑guided gate learns to route samples to low‑uncertainty experts.
	
	Despite these advances, existing uncertainty-aware MoE methods for MSA such as EUAR~\cite{gao2024enhanced} still adopt a symmetric expert design, where all modalities are treated as equally important peers. Their routing decisions are driven primarily by uncertainty estimates or polarity patterns, without explicitly modeling the inherent asymmetry between the dominant textual modality and the auxiliary non-verbal modalities. Moreover, these methods either estimate uncertainty separately from the routing network or rely on post-hoc adjustments, rather than integrating per-sample, per-modality predictive uncertainty directly into the end-to-end router logits. Separately, MMoLRE~\cite{zhang2025multimodal} demonstrates the benefit of separating shared and task-specific experts in multi-task scenarios, but its design is not tailored to the modality-asymmetry challenge in single-task MSA.

	To address these gaps, TAME introduces an asymmetric shared-plus-routed expert organization, where a shared text expert is always active and routed cross-modal experts are sparsely activated. Crucially, we introduce Prophet Heads to produce per-sample uncertainty estimates for each modality, and these estimates are directly injected into the router logits and trained jointly with the entire architecture. This allows the model to learn, end-to-end, when a non-verbal modality is trustworthy enough to complement the text, thereby unifying asymmetric modality fusion with reliability‑aware routing.
	
	\section{Methodology}
	\label{sec:method}
	
\begin{figure*}[t]
		\centering
		\includegraphics[width=\textwidth]{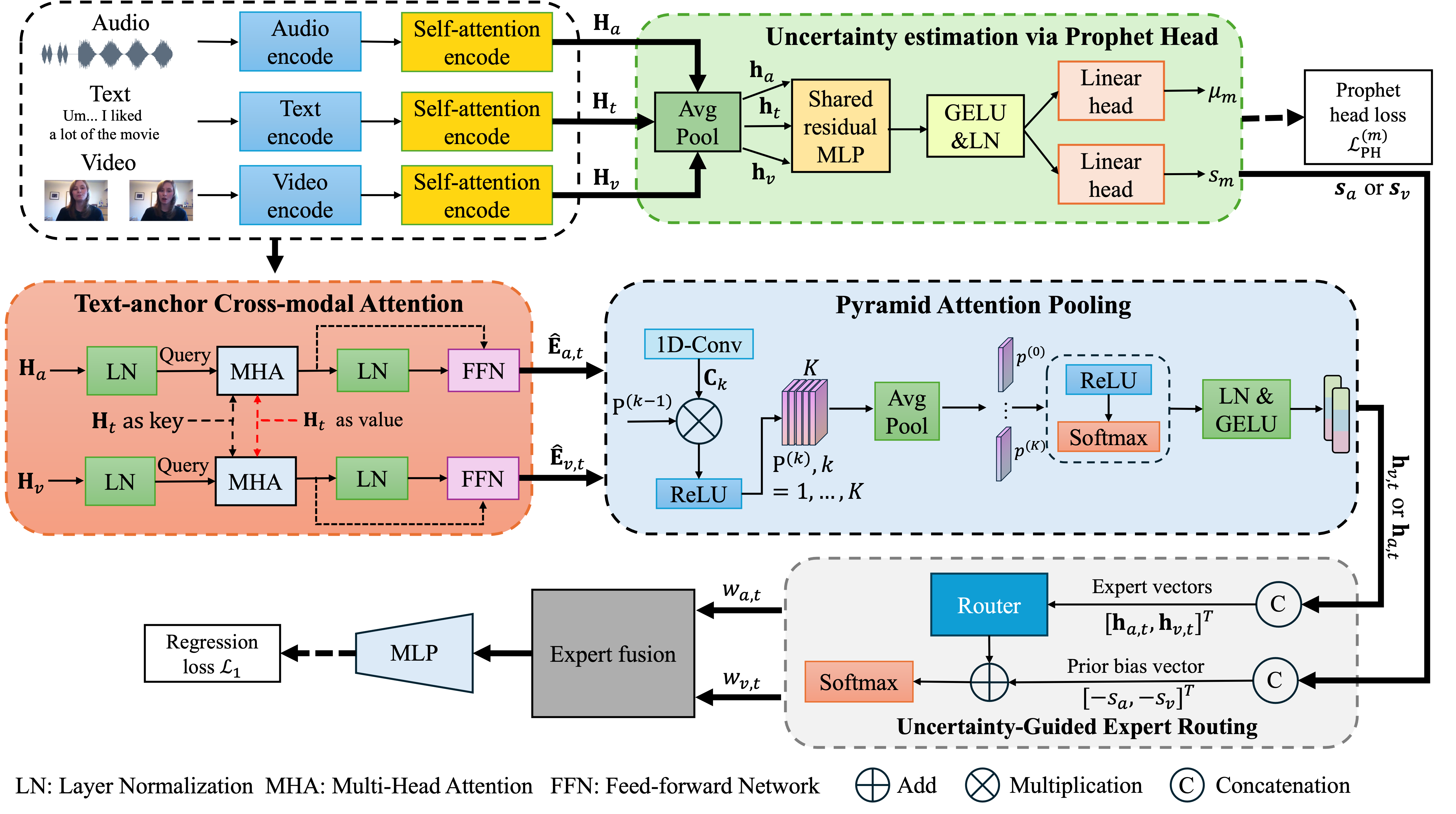}
		\caption{Overview of the TAME architecture. It consists of a Text-Anchored Cross-Modal Attention (TACMA), a Pyramid Attention Pooling (PAP), Uncertainty Estimation via Prophet Heads (UEPH), and an Uncertainty-Guided Expert Router (UGER).}
		\label{fig:overview}
	\end{figure*}

	TAME formulates this task as an uncertainty-guided refinement paradigm based on a language prior, where nonverbal evidence is selectively leveraged based on its reliability. The proposed TAME framework, as shown in Fig.~\ref{fig:overview}, consists of four core components: (1) a Text-Anchored Cross-Modal Attention (TACMA) for interpreting nonverbal signals through the lens of linguistic semantics, (2) a Pyramid Attention Pooling (PAP) for capturing multi-scale temporal structure, (3) Uncertainty Estimation via Prophet Heads (UEPH) for quantifying the predictive uncertainty of each modality-specific expert, and (4) an Uncertainty-Guided Expert Router (UGER) for adaptively determining the weight of corrective evidence from experts.
	
	\subsection{Problem Formulation}
	\label{ssec:formulation}
	
	Given an utterance with multimodal observations, let $\mathbf{X}_t \in \mathbb{R}^{T_t \times d_t}$, $\mathbf{X}_a \in \mathbb{R}^{T_a \times d_a}$, and $\mathbf{X}_v \in \mathbb{R}^{T_v \times d_v}$ denote the text, audio, and video feature sequences, respectively, where $T_m$ and $d_m$ are the temporal length and feature dimensionality for modality $m \in \{t, a, v\}$, respectively.
	The objective of MSA is to predict a continuous sentiment intensity $y$ by extracting and integrating complementary information across these modalities.
	
	Traditional multimodal fusion strategies typically treat all modalities as symmetric contributors and learns a single joint representation $\mathbf{h} = f(\mathbf{X}_t, \mathbf{X}_a, \mathbf{X}_v)$. However, this symmetric design fails to account for the inherent asymmetry in the informativeness of different modalities. Recently, the shared-plus-routed expert paradigm proposed in DeepSeekMoE~\cite{dai2024deepseekmoe} employs a subset of always-activate shared experts to capture common knowledge, alongside sparsely activated routed experts dedicated to specialized processing. 
	
	Motivated by this paradigm, we propose a TAME architecture to decompose the prediction process into a shared expert and two routed experts, which explicitly models the empirically observed asymmetry in modality informativeness. Specifically, the text branch serves as an always-on shared expert providing a stable semantic prior $\mathbf{h}_t$, while
	the audio-text and video-text cross-modal branches act as routed experts, supplying corrective evidence denoted as
	$\mathbf{h}_{a,t}$ and $\mathbf{h}_{v,t}$, respectively. Critically, their contributions are dynamically modulated by a learned uncertainty-aware router, leading to the following prediction function:
	\begin{equation}
		\hat{y}=f(\underbrace{\mathbf{h}_t}_{\text{shared prior}}+
		\underbrace{w_{a,t}\,\mathbf{h}_{a,t} + w_{v,t}\,\mathbf{h}_{v,t}}_{\text{routed corrective evidence}}),
		\label{eq:overview}
	\end{equation}
	where $w_{a,t}, w_{v,t} \ge 0$ with $w_{a,t} + w_{v,t} = 1$ denote sample-level routing weights,
	and $f$ denotes an learned prediction function of our TAME model.
	
	\subsection{Modality Encoding}
	\label{ssec:encoding}

As the temporal lengths differ across modalities,~\ie, $T_a \ne T_v \ne T_t$, a linear temporal alignment layer is employed to map each non-text modality to the fixed length~$T$ before encoding, that is, $\mathbf{X}'_m = \mathbf{X}_m \mathbf{W}^{m}_{\mathrm{align}} \in \mathbb{R}^{T \times d_m}$, where $\mathbf{W}^{m}_{\mathrm{align}} \in \mathbb{R}^{T_m \times T}$ is a learned alignment matrix. This temporal alignment operates in the feature space and does not require external word-level alignment tools. 

\textbf{Text encoding.} Raw token sequences from the dataset are fed directly into the model and encoded end-to-end by a fine-tuned DeBERTa-v3-base~\cite{he2023debertav} for the two English benchmarks,~\ie, CMH-MOSI and CMU-MOSEI. For CH-SIMS, the corresponding Chinese pre-trained language model, such as BERT-base-Chinese~\cite{devlin2019bert} or ERNIE-3.0-base~\cite{sun2021ernie3}, is fine-tuned in the same end-to-end fashion. The output consists of 768-dimensional contextualized token embeddings. We denote these text representations as $\mathbf{Z}_t \in \mathbb{R}^{T \times d_t}$, where $d_t = 768$. 

\textbf{Audio/Video encoding.} Acoustic features $\mathbf{Z}_a$ are the pre-extracted COVAREP descriptors released with each dataset ($d_a=5$ for CMU-MOSI; $d_a=74$ for CMU-MOSEI), and visual features $\mathbf{Z}_v$ are the pre-extracted Facet facial-expression descriptors ($d_v=20$ for CMU-MOSI; $d_v=35$ for CMU-MOSEI). For CH-SIMS we use the official multimodal features released with the dataset~\cite{yu2020chsims} for both modalities.
	
\textbf{Self-attention encoding.} After projection, each modality sequence is further contextualized through an independent $L$-layer Transformer encoder~\cite{vaswani2017attention},~\ie, $ \mathbf{H}_m=\mathcal{T}_m(\mathbf{Z}_m) \in \mathbb{R}^{T \times d}$, where  $m \in \{a, v, t\}$, $\mathcal{T}_m$ denotes the modality-specific Transformer encoder. Each $\mathcal{T}_m$ consists of multi-head self-attention with $h$~heads, Pre-LN residual connections, and $\GELU$-activated feed-forward networks.
	
	\subsection{Text-Anchored Cross-Modal Attention}
	\label{ssec:cross_attn}
	
	According to the prior studies~\cite{wharton2024non}, nonverbal features are often interpreted together with the linguistic context. Meanwhile, in text-centric benchmarks, text has been shown to provide the strongest and most stable semantic signal~\cite{yu2021learning, das2023multimodal}. Using text as the contextual reference can ensure that cross-modal representations are firmly grounded in reliable semantic information, avoiding misinterpretation of nonverbal cues divorced from linguistic context. Motivated by these observations, we propose the text-anchored cross-modal attention (TACMA) mechanism to anchor the interpretation of nonverbal features to text.  Concretely, for each non-text modality $m \in \{a, v\}$, a Pre-LN cross-attention layer uses $\mathbf{H}_m$ as queries and $\mathbf{H}_t$ as keys and values. The formulation is defined as follows,
	\begin{align}
		\mathbf{E}_{m,t} &= \mathbf{H}_m + \MHA\!\big(\LN(\mathbf{H}_m),\, \mathbf{H}_t,\, \mathbf{H}_t\big),
		\label{eq:ca_attn}\\
		\hat{\mathbf{E}}_{m,t} &= \mathbf{E}_{m,t} + \FFN\!\big(\LN(\mathbf{E}_{m,t})\big),
		\label{eq:ca_ffn}
	\end{align}
	where $\MHA(\cdot)$ denotes multi-head attention~\cite{vaswani2017attention} with $h$~heads, $\LN(\cdot)$ denotes layer normalization, and $\FFN(\cdot)$ is a two-layer feed-forward network with $\GELU$ activation. 
	
	This yields the cross-modal expert sequences
	$\hat{\mathbf{E}}_{a,t} \in \mathbb{R}^{T \times d_{a,t}}$ and
	$\hat{\mathbf{E}}_{v,t} \in \mathbb{R}^{T \times d_{v,t}}$, respectively.
	Notably, TAME restricts cross-attention exclusively to the unidirectional A$\to$T and V$\to$T pathways, and deliberately excludes direct A$\leftrightarrow$V interactions. This design is not merely a structural simplification, but a principled choice motivated by robustness and semantic consistency. Direct audio-video cross-attention may create an information pathway that bypasses the text anchor, allowing spurious correlations between nonverbal cues such as lip movements correlating with speech tempo to introduce noise and distort the routing signal. By anchoring all cross-modal fusion to text, we enforce that information flows through a linguistic bottleneck, which serves as a semantic filter to ensure reliable and interpretable multimodal reasoning.
	
	\subsection{Pyramid Attention Pooling}
	\label{ssec:pap}
	
	Since each cross-modal expert sequence $\hat{\mathbf{E}}_{m,t} \in \mathbb{R}^{T \times d_{m,t}}$ has variable temporal length, it must be condensed into a fixed-dimensional representation for routing and fusion.
	A naive solution employs simple temporal average pooling, which treats all time steps equally and discards multi-scale temporal structure. However, sentiment-relevant cues manifest at different temporal granularities. These include fine-grained word-level emphasis, medium-grained phrase-level prosody, and coarse-grained utterance-level affect. Therefore, we propose Pyramid Attention Pooling (PAP), which captures
	multi-scale temporal structure via a learnable temporal pyramid with adaptive scale weighting. This module comprises three key components including pyramid construction, adaptive scale attention, and output projection.
	
	\textbf{Pyramid construction.} Given input $\hat{\mathbf{E}}_{m,t}$, PAP constructs a hierarchy of $K$ resolution levels,~\ie, $\mathbf{P}^{(0)} = \hat{\mathbf{E}}_{m,t}$. Level~0 retains the original sequence. Each subsequent level halves the temporal resolution through a learnable stride-2 convolution with ReLU activation:
	\begin{equation}
		\mathbf{P}^{(k)} = \ReLU\!\big(\mathbf{C}_{k} \ast \mathbf{P}^{(k-1)}\big),
		\; k = 1, \dots, K{-}1,
		\label{eq:pap_pyramid}
	\end{equation}
	where $\mathbf{C}_{k}$ denotes a 1D convolution with stride 2 and kernel size~2 and $d$~output channels. Different from average pooling used in FPN~\cite{lin2017feature}, the learnable convolution enables each pyramid level to selectively preserve temporally salient features while suppressing noise at the corresponding scale. Each level is then globally average-pooled to obtain a scale descriptor,~\ie,
	$\mathbf{p}^{(k)}= \frac{1}{T_k}\sum_{i=1}^{T_k}\mathbf{P}^{(k)}_i$.
	
	\textbf{Adaptive scale attention.}
	The relative importance of different temporal scales varies across samples. A sample with a single emphatic word benefits more from fine-grained (\ie, level~0) information, while a sample with gradually shifting affect depends on coarser summaries. Motivated by the channel attention mechanism in SE-Net~\cite{hu2018squeeze}, we employ a lightweight two-layer network to compute scale-specific weights as follows:
	\begin{equation}
		\boldsymbol{\alpha} = \softmax\!\Big(
		\mathbf{W}_2\,\ReLU\!\big(\mathbf{W}_1 [\mathbf{p}^{(0)}; \dots; \mathbf{p}^{(K-1)}]\big)
		\Big),
		\label{eq:pap_weights}
	\end{equation}
	where $\mathbf{W}_1 \in \mathbb{R}^{d \times (K\times d)}$ and $\mathbf{W}_2 \in \mathbb{R}^{K \times d}$ are learnable weights of two-layer bottleneck that forces the network to attend to the most discriminative scale characteristics.
	The softmax normalization ensures $\sum_k \alpha_k = 1$, yielding a convex combination of scale descriptors.
	
	\textbf{Output projection.}
	The pooled representation is obtained by the weighted sum of scale descriptors, followed by a linear projection with layer normalization:
	\begin{equation}
		\mathbf{h}_{m,t} = \sigma\!\bigl(\LN\bigl(\mathbf{W}_p \textstyle\sum_{k=0}^{K-1} \alpha_k \, \mathbf{p}^{(k)}\bigr)\bigr),
		\label{eq:pap_output}
	\end{equation}
	where $\sigma(\cdot)$ denotes dropout followed by $\GELU$ activation.
	
	In the implementation, PAP is applied independently to each routed expert branch, producing $\mathbf{h}_{a,t}$ and $\mathbf{h}_{v,t}$ as input to both the router and the fusion stage. The shared text expert uses simple temporal average pooling to otbain $\mathbf{h}_t$, since text features from BERT are already well-contextualized at the token level and do not exhibit the same multi-scale temporal variation as audio-visual signals.
	
	\subsection{Uncertainty Estimation via Prophet Heads}
	\label{ssec:prophet}
	
	Before routing, TAME independently assesses the reliability of each modality for the current sample. Therefore, we introduce a lightweight Prophet Head (PH) strategy for each modality branch after self-attention encoding to quantify modality reliability. Specifically, we model the conditional distribution of the sentiment label given modality-$m$ features as a Gaussian with input-dependent variance, denoted as $p(y|\mathbf{H}_m) = \mathcal{N}(y;\, \mu_m,\, \sigma_m^2)$. Here, $\mu_m$ corresponds to the predicted sentiment score,  and $\sigma_m^2$ represents the aleatoric uncertainty. Both values are estimated from the learned feature representation. 
	
	Each Prophet head takes the average pooled self-attention output $\mathbf{h}_m = \frac{1}{T}\sum_{i}\mathbf{H}_m^{(i)}$ as input. A residual MLP with Post-LN serves as a feature extractor to produce a compact and discriminative representation:
	\begin{equation}
		\mathbf{f}_m = \LN\!\big(\bar{\mathbf{W}}_2\sigma(\bar{\mathbf{W}}_1 \mathbf{h}_m) + \mathbf{h}_m\big),
		\label{eq:prophet_feat}
	\end{equation}
	where $\bar{\mathbf{W}}_1 \in \mathbb{R}^{2d \times d}$ and $\bar{\mathbf{W}}_2 \in \mathbb{R}^{d \times 2d}$ are learnable parameters, and $\sigma(\cdot)$ denotes $\GELU$ activation followed by dropout.  
	
	Based on this enhanced shared feature, two separate linear heads predict the sentiment mean $ \mu_m= \mathbf{w}_\mu^\top \mathbf{f}_m + b_\mu, \label{eq:prophet_mu}$ and the log-variance $s_m = \mathbf{w}_s^\top \,\mathbf{f}_m + b_s$, where $\mathbf{w}_\mu$ and $\mathbf{w}_s$ denote the learnable weights. 
	
	This heteroscedastic modeling strategy enables the system to capture sample-dependent noise levels. A small value of  $\sigma_m^2$ indicates that the modality provides clear and trustworthy sentiment evidence, while a large value suggests that the signal is ambiguous, noisy, or otherwise unreliable. Importantly, aleatoric uncertainty describes the irreducible noise that exists naturally in the data. As proper initialization of the PH is essential to avoid training instability, especially in the early stages, we initialize the variance head with $b_s=0$ and $\mathbf{w}_s \sim \mathcal{N}(0, 0.01^2)$, a setting that gives $\sigma_m^2 \approx e^{0} = 1$ at initialization. 
	
	To train each head to produce calibrated uncertainty estimates, we define a Prophet head loss based on heteroscedastic Gaussian negative log-likelihood (GNLL) as the supervision signal:
	\begin{equation}
		\mathcal{L}_{\text{PH}}^{(m)}
		= \frac{1}{2}\,e^{-s_m}(y - \mu_m)^2 + \frac{1}{2}\,s_m.
		\label{eq:nll}
	\end{equation}
	
	This loss function encourages accurate predictions and well-calibrated uncertainty by balancing two key terms. The first term is a precision-weighted regression loss that strengthens training for confident predictions, while the second term acts as a regularization penalty on log-variance to prevent the model from concealing errors by setting overly large $s_m$. The balance between these two terms yields meaningful and calibrated uncertainty. Uniformly high $s_m$ reduces the weighted error but increases the penalty, while uniformly low $s_m$ forces the model to fit noisy samples and degrades overall performance. An optimal equilibrium emerges when $s_m$ faithfully represents the true difficulty of each prediction.
	
	\subsection{Uncertainty-Guided Expert Routing}
	\label{ssec:router}
	
	The router determines how much corrective evidence from each routed expert should be admitted  into the final prediction. Its design combines content-based gating and an uncertainty prior. The gating reflects what the cross-modal features themselves suggest about their relevance, while the prior indicates how reliable each expert's prediction is likely to be.
	
	Given expert representations $\mathbf{h}_{a,t}$ and $\mathbf{h}_{v,t}$ from PAP, the router first concatenates them and computes content-based gating logits through a bottleneck MLP:
	\begin{equation}
		\mathbf{g} = \mathbf{W}_g^{(2)}\,\mathrm{Norm}\!\big(
		\GELU(\mathbf{W}_g^{(1)} [\mathbf{h}_{a,t}, \mathbf{h}_{v,t}])
		\big),
		\label{eq:router_content}
	\end{equation}
	where $\mathbf{W}_g^{(1)} \in \mathbb{R}^{d/4 \times 2d}$ and $\mathbf{W}_g^{(2)} \in \mathbb{R}^{2 \times d/4}$ are the weights of bottleneck, and $\mathrm{Norm}(\cdot)$ denotes $\ell_2$-normalization.
	
	This architecture servers two key purposes. First, it reduces the parameter complexity of the router, preventing it to specific sample patterns and ensuring the model focuses on generalizable feature relationship.  Second,  the $\ell_2$-normalization stabilizes the scale of the gating logits across different samples, avoding bias caused by varying feature magnitudes.
	
	Given  a prior bias vector $\mathbf{b} = [-s_a, -s_v]^\top$ outputted by the Prophet Heads, a lower $s_m$ corresponds to a larger prior that encourages the corresponding expert. The content-based logits and the uncertainty prior are combined through a learnable weighting:
	\begin{equation}
		[w_{a,t},\, w_{v,t}] = \softmax\!\bigg(
		\frac{\mathbf{g} + \lambda_p \cdot \mathbf{b}}{\tau}
		\bigg),
		\label{eq:router_final}
	\end{equation}
	where $\tau$ is a temperature parameter that controls the sharpness of the routing distribution, and  the prior strength $\lambda_p$ is learnable scale controlling the strengthen of the uncertainty prior.  The temperature $\tau$ controls the entropy of the routing distribution. When $\tau \to 0$, routing degenerates to hard expert selection. In contrast, when $\tau \to \infty$, routing becomes uniform,~\ie, $w_{a,t} = w_{v,t} = 0.5$. In our implementation, according to our empirical experience, we fix $\tau = 1.0$. 
	
	\subsection{Fusion and Training Objective}
	\label{ssec:fusion}
	
	\textbf{Expert fusion.} The shared expert representation of text modality is obtained by average pooling the text self-attention output,~\ie, $\mathbf{h}_t = \frac{1}{T}\sum_i \mathbf{H}_t^{(i)}$. The final fused representation combines the shared prior with the routed corrective evidence as in Equation~\eqref{eq:overview}:
	\begin{equation}
		\mathbf{c} = \mathbf{h}_t + w_{a,t}\,\mathbf{h}_{a,t} + w_{vt}\,\mathbf{h}_{v,t}.
		\label{eq:fusion}
	\end{equation}
	
	In Equation~\eqref{eq:fusion}, $\mathbf{h}_t$ serves as the prior prediction, and the routed term provides a perturbation whose direction and magnitude are controlled by the router. This is structurally analogous to a residual connection~\cite{he2016deep}, where the routed experts learn to predict the residual sentiment information that the text alone cannot capture.
	
	\textbf{Output projection.} A residual two-layer MLP produces the final sentiment prediction:
	\begin{align}
		\mathbf{c}' &= \ReLU(\mathbf{W}_1^{o}\, \mathbf{c} + \mathbf{b}_1^{o}), \label{eq:out_mlp1}\\
		\hat{y} &= \mathbf{w}_o^\top (\mathbf{W}_2^{o}\, \mathbf{c}' + \mathbf{c}) + b_o, \label{eq:out_mlp2}
	\end{align}
	where dropout is applied after the $\ReLU$ activation. The skip connection from $\mathbf{c}$ to the output layer ensures that the fused representation is directly accessible to the prediction head, preventing information loss through the nonlinear projection.
	
	\textbf{Training objective.} Based on Equation~\eqref{eq:nll}, the total loss combines the main regression loss and the
	auxiliary Prophet Head losses, which is formulated as follows:
	\begin{equation}
		\mathcal{L} = \mathcal{L}_1(\hat{y}, y)
		+ \frac{1}{3}\sum_{m \in \{t,a,v\}}
		\mathcal{L}_{\text{PH}}^{(m)},
		\label{eq:loss}
	\end{equation}
	where $\mathcal{L}_1= |\hat{y} - y|$ denotes the $\ell_1$ loss. Here, we choose the $\ell_1$ instead of $\ell_2$ loss because it is more robust to outlier sentiment annotations~\cite{zadeh2016mosi}. The Prophet head losses serve two key purposes besides uncertainty estimation. First, they train the uncertainty heads to generate calibrated $s_m$ values for routing. Second, they provide auxiliary gradients to each modality encoder, encouraging these encoders to learn representations that are individually predictive of the sentiment label. 
	
	\section{Experimental Settings}
	\label{ssec:settings}
	
	\subsection{Datasets and implementation details}
	\label{ssec:datasets}
	
	We conduct experiments on three publicly available multimodal sentiment analysis benchmarks: CMU-MOSI~\cite{zadeh2016mosi}, CMU-MOSEI~\cite{zadeh2018multimodal}, and CH-SIMS~\cite{yu2020chsims}. CMU-MOSI contains 2,199 video clips annotated with continuous sentiment score from $[-3, +3]$. We use the official train/validation/test split of 1,284/229/686 utterances. CMU-MOSE is a larger-scale multimodal sentiment dataset with 23,453 utterances  on the same $[-3, +3]$ scale. The official split consists of  16,326/1,871/4,659 utterances for training, validation, and testing. CH-SIMS is a Chinese-language dataset comprising 2,281 video segments annotated with fine-grained sentiment intensity within $[-1, +1]$. We follow the official train/validation/test split of 1,368/456/457 segments. 

	TAME employs eight attention heads for both self-attention and cross-attention modules. The number of self-attention layer is set to 3, 3, 4, 3, and 2 layers for CMU-MOSI (Aligned), CMU-MOSI (Unaligned), CMU-MOSEI (Aligned), CMU-MOSEI (Unaligned), and CH-SIMS, respectively; the corresponding pyramid levels for PAP are 2, 2, 4, 4, and 3. All Prophet Heads adopt a hidden expansion ratio of~$2\times$ with GELU activation, and the router temperature $\tau$ is fixed to 1.  We optimize with AdamW~\cite{loshchilov2019decoupled} on two NVIDIA V100 GPUs with 32GB each, using a batch size of 16, gradient clipping at a threshold of 0.6, and a ReduceLROnPlateau scheduler at a patience of 5 and a factor of 0.5. Early stopping is applied. All experiments are implemented in PyTorch.
	
	For CMU-MOSI and CMU-MOSEI, following prior work~\cite{fang2025emoe,zhao2025sdrs}, we report Acc-2 (negative/positive), Acc-7, F1 score, and MAE under both aligned and unaligned settings. For CH-SIMS, following the official  protocol~\cite{yu2020chsims}, we report Acc-2 (negative vs.\ non-negative), Acc-3, Acc-5, F1, MAE, and Pearson correlation (Corr). Except for MAE, higher value denote better performance.
	
	\subsection{Comparison with State-of-the-Art}
	\label{ssec:sota}
	
		\begin{table*}[t!]
		\centering
		\caption{Comparison with state-of-the-art methods on CMU-MOSI and CMU-MOSEI.
			Best results are in \textbf{bold}; second best are \underline{underlined}.
			$^\ast$~indicates vanilla BERT-base language features. $^\S$~indicates a domain-adapted BERT variant continued-pretrained on MSA data (PAMoE's MSA-BERT). $^\ddagger$~indicates RoBERTa-based language features. $^\dagger$~indicates DeBERTa-v3-base language features. The symbol `--' indicates `Not reported'.}
		\label{tab:mosi_mosei_combined}
		\footnotesize
		\setlength{\tabcolsep}{3pt} 
		\begin{tabular}{
				p{2.2cm}  
				>{\centering\arraybackslash}p{0.8cm}
				>{\centering\arraybackslash}p{0.8cm}
				>{\centering\arraybackslash}p{0.8cm}
				>{\centering\arraybackslash}p{0.8cm}
				>{\centering\arraybackslash}p{0.8cm}
				>{\centering\arraybackslash}p{0.8cm}
				>{\centering\arraybackslash}p{0.8cm}
				>{\centering\arraybackslash}p{0.8cm}
				>{\centering\arraybackslash}p{0.8cm}
				>{\centering\arraybackslash}p{0.8cm}
				>{\centering\arraybackslash}p{0.8cm}
				>{\centering\arraybackslash}p{0.8cm}
				>{\centering\arraybackslash}p{0.8cm}
				>{\centering\arraybackslash}p{0.8cm}
				>{\centering\arraybackslash}p{0.8cm}
				>{\centering\arraybackslash}p{0.8cm}
			}
			\toprule
			& \multicolumn{8}{c}{\textbf{CMU-MOSI}} & \multicolumn{8}{c}{\textbf{CMU-MOSEI}} \\
			\cmidrule(lr){2-9} \cmidrule(lr){10-17}
			& \multicolumn{4}{c}{\textbf{Aligned}} & \multicolumn{4}{c}{\textbf{Unaligned}} & \multicolumn{4}{c}{\textbf{Aligned}} & \multicolumn{4}{c}{\textbf{Unaligned}} \\
			\cmidrule(lr){2-5} \cmidrule(lr){6-9} \cmidrule(lr){10-13} \cmidrule(lr){14-17}
			\textbf{Method} & Acc2 (\%) & F1 (\%) & Acc7 (\%) & MAE & Acc2 (\%) & F1 (\%) & Acc7 (\%) & MAE & Acc2 (\%) & F1 (\%) & Acc7 (\%) & MAE & Acc2 (\%) & F1 (\%) & Acc7 (\%) & MAE \\
			\midrule

				MISA$^\ast$~\cite{hazarika2020misa}
			& 84.2 & 84.2 & 41.8 & 0.754 & 83.8 & 83.9 & 43.6 & 0.742
			& 85.3 & 85.1 & 52.3 & 0.543 & 84.8 & 84.8 & 51.0 & 0.557 \\
				Self-MM$^\ast$~\cite{yu2021learning}
			& 84.9 & 84.9 & 45.3 & 0.738 & 83.4 & 83.6 & 45.7 & 0.724
			& 84.5 & 84.3 & 53.2 & 0.540 & 85.3 & 84.8 & 52.9 & 0.535 \\
			
				TFN~\cite{zadeh2017tensor}
			& 78.8 & 78.9 & 31.9 & 0.953 & 76.5 & 76.6 & 35.3 & 0.995
			& 80.4 & 80.7 & 50.9 & 0.574 & 84.2 & 84.0 & 50.2 & 0.573 \\
				LMF~\cite{liu2018efficient}
			& 78.7 & 78.7 & 36.9 & 0.931 & 79.1 & 79.1 & 31.1 & 0.963
			& 84.7 & 84.5 & 52.3 & 0.564 & 83.8 & 83.9 & 51.9 & 0.565 \\
				MulT~\cite{tsai2019multimodal}
			& 80.0 & 80.1 & 35.1 & 0.936 & 80.3 & 80.3 & 33.2 & 0.933
			& 82.7 & 82.8 & 52.3 & 0.572 & 84.0 & 84.0 & 53.2 & 0.556 \\
				RAVEN~\cite{wang2019words}
			& 78.0 & 76.6 & 33.2 & 0.915 & 72.7 & 73.1 & 31.7 & 1.076
			& 79.1 & 79.5 & 50.0 & 0.614 & 75.4 & 75.7 & 45.5 & 0.664 \\
			\midrule
			
			EF-LSTM~\cite{williams2018recognizing}
			& 75.3 & 75.2 & 33.7 & 1.386 & 73.6 & 74.5 & 31.0 & 1.420
			& 78.2 & 77.9 & 47.4 & 0.620 & 76.1 & 75.9 & 46.3 & 0.594 \\
			LF-DNN~\cite{williams2018recognizing}
			& 78.4 & 78.3 & 31.5 & 0.972 & 78.2 & 78.3 & 32.5 & 0.987
			& 83.5 & 83.1 & 51.7 & 0.568 & 83.7 & 83.2 & 52.3 & 0.561 \\
				FDMER$^\ast$~\cite{yang2022disentangled}
			& 84.6 & 84.7 & 44.1 & 0.724 & -- & -- & -- & --
			& 86.1 & 85.8 & 54.1 & 0.536 & -- & -- & -- & -- \\
			MMIM$^\ast$~\cite{han2021improving}
		& 84.6 & 84.5 & 45.8 & 0.717 & 83.4 & 83.4 & 45.9 & 0.777
		& 83.6 & 83.5 & 50.1 & 0.580 & 81.5 & 81.3 & 52.6 & 0.578 \\
		\midrule
		
			EUAR$^\ast$~\cite{gao2024enhanced}
		& 86.3 & 86.3 & 46.1 & -- & -- & -- & -- & --
		& 86.6 & 86.4 & 54.9 & -- & -- & -- & -- & -- \\
					U-MOE$^\ast$~\cite{xiao2026uncertainty}
		& 79.51 & 80.27 & 42.98 & 0.808 & -- & -- & -- & --
		& 84.14 & 84.02 & 50.66 & 0.572 & -- & -- & -- & -- \\
			PAMoE$^\S$~\cite{huang2025pamoe}
		& 87.04 & 86.99 & \underline{48.7} & \textbf{0.609} & -- & -- & -- & --
		& \underline{87.07} & \underline{86.90} & 54.6 & 0.526 & -- & -- & -- & -- \\
		MMoLRE$^\ddagger$~\cite{zhang2025multimodal}
		& \underline{88.21} & \underline{88.23} & 47.52 & 0.666 & -- & -- & -- & --
		& 86.57 & 86.28 & \underline{55.78} & \underline{0.505} & -- & -- & -- & -- \\
					EMOE$^\ast$~\cite{fang2025emoe}
		& 85.4 & 85.4 & 47.7 & 0.710 & \underline{85.4} & \underline{85.3} & \underline{47.8} & \underline{0.697}
		& 85.3 & 85.3 & 54.1 & 0.536 & \underline{85.5} & \underline{85.5} & \underline{53.9} & \underline{0.530} \\
		\midrule
		
			PMR~\cite{lv2021progressive}
		& 83.6 & 83.4 & 40.6 & -- & 82.4 & 82.1 & 40.6 & --
		& 83.3 & 82.6 & 52.5 & -- & 83.1 & 82.8 & 51.8 & -- \\
		MICA~\cite{liang2021attention}
		& -- & -- & -- & -- & 82.6 & 82.7 & 40.8 & --
		& 83.7 & 83.3 & 52.4 & -- & -- & -- & -- & -- \\
			IMDer$^\ast$~\cite{wang2023incomplete}
		& 85.7 & 85.6 & 45.3 & 0.736 & 85.2 & 85.2 & 44.5 & 0.744
		& 85.1 & 85.1 & 53.4 & 0.542 & 84.7 & 84.7 & 53.0 & 0.582 \\
		GCNet$^\ast$~\cite{lian2023gcnet}
		& 85.2 & 85.1 & 44.9 & -- & -- & -- & -- & --
		& 85.2 & 85.1 & 51.5 & -- & -- & -- & -- & -- \\
			TCTR$^\ast$~\cite{yang2026tctr}
		& 86.1 & 86.3 & 44.5 & 0.751 & -- & -- & -- & --
		& 85.8 & 85.8 & 53.9 & 0.532 & -- & -- & -- & -- \\
			DLF$^\ast$~\cite{wang2025dlf}
		& 85.1 & 85.0 & 47.1 & 0.731 & -- & -- & -- & --
		& 85.4 & 85.3 & 53.9 & 0.536 & -- & -- & -- & -- \\
			DEVA$^\ast$~\cite{wu2025enriching}
		& 86.3 & 86.3 & 46.3 & 0.730 & -- & -- & -- & --
		& 86.1 & 86.2 & 52.3 & 0.541 & -- & -- & -- & -- \\
		
			MFN~\cite{zadeh2018memory}
			& 78.4 & 78.4 & 35.6 & 0.964 & 80.0 & 80.1 & 34.7 & 0.971
			& 84.0 & 84.0 & 50.8 & 0.574 & 83.2 & 83.3 & 51.3 & 0.567 \\
			Graph-MFN~\cite{zadeh2018multimodal}
			& 78.1 & 78.1 & 31.5 & 0.970 & 79.4 & 79.2 & 34.4 & 0.930
			& 84.6 & 84.5 & 51.6 & 0.553 & 84.2 & 84.2 & 51.8 & 0.568 \\
			MCTN~\cite{pham2019found}
			& 79.3 & 80.0 & 33.1 & 0.963 & 77.1 & 77.3 & 31.9 & 1.033
			& 83.5 & 83.4 & 49.4 & 0.590 & 82.3 & 82.5 & 52.0 & 0.572 \\
		
			DMD$^\ast$~\cite{li2023decoupled}
		& 83.2 & 83.2 & 46.2 & 0.721 & 84.0 & 84.0 & 46.7 & 0.721
		& 84.8 & 84.7 & 52.4 & 0.546 & 84.7 & 84.7 & 53.1 & 0.536 \\

			\midrule
			Ours $^\dagger$
			& \textbf{88.41} & \textbf{88.41} & \textbf{49.42} & \underline{0.650} & \textbf{88.41} & \textbf{88.40} & \textbf{49.13} & \textbf{0.674}
			& \textbf{88.25} & \textbf{88.23} & \textbf{56.01} & \textbf{0.499} & \textbf{88.22} & \textbf{88.15} & \textbf{56.49} & \textbf{0.497} \\
			\bottomrule
		\end{tabular}
	\end{table*}

\begin{table}[th!]
	\centering
	\caption{Comparison with state-of-the-art methods on CH-SIMS. Best results are in \textbf{bold}; second best are \underline{underlined}. \texttt{[N]}/\texttt{[U]} indicate that the model is trained \emph{without} or \emph{with} additional unimodal sentiment labels, respectively. \texttt{(B)}/\texttt{(E)} indicate that the textual embedding is extracted by BERT-base-Chinese or ERNIE-3.0-base, respectively. The asterisk ($\ast$) on baseline names follows the original SDRS convention denoting an improved fine-tuned variant.}
	\label{tab:chsims}
	\footnotesize
	\setlength{\tabcolsep}{3pt}
	\begin{tabular}{lcccccc}
		\toprule
		\textbf{Method} & \makecell{Acc-2 \\(\%)} & \makecell{F1 \\(\%)} & \makecell{Acc-3 \\(\%)} & \makecell{Acc-5 \\ (\%)} & MAE & Corr \\
		\midrule
		LF-DNN\,\texttt{[N]}~\cite{williams2018recognizing}    & 78.87 & 79.87 & 66.91 & 41.62 & 0.420 & 0.612 \\
		MFN\,\texttt{[N]}~\cite{zadeh2018memory}       & 77.90 & 77.88 & 65.73 & 39.47 & 0.435 & 0.582 \\
		LMF\,\texttt{[N]}~\cite{liu2018efficient}       & 77.77 & 77.88 & 64.68 & 40.53 & 0.441 & 0.576 \\
		TFN\,\texttt{[N]}~\cite{zadeh2017tensor}       & 78.38 & 78.62 & 65.12 & 39.30 & 0.432 & 0.591 \\
		Self-MM\,\texttt{[N]}~\cite{yu2021learning}   & 80.04 & 80.44 & 65.47 & 41.53 & 0.425 & 0.595 \\
		LF-DNN\,\texttt{[U]}~\cite{williams2018recognizing}    & 82.28 & 82.52 & 69.06 & 38.03 & 0.406 & 0.674 \\
		LMF\,\texttt{[U]}~\cite{liu2018efficient}       & 82.32 & 81.84 & 67.70 & 37.33 & 0.420 & 0.631 \\
		TFN\,\texttt{[U]}~\cite{zadeh2017tensor}       & 82.45 & 82.56 & 69.02 & 37.20 & 0.406 & 0.669 \\
		Self-MM$^\ast$\,\texttt{(U)(B)}~\cite{yu2021learning} & 81.32 & 81.73 & 68.39 & 39.77 & 0.408 & 0.647 \\
		ConFEDE\,\texttt{(U)(B)}~\cite{yang2023confede}        & 82.23 & 82.08 & 70.15 & 46.30 & 0.392 & 0.637 \\
		SDRS\,\texttt{(N)(B)}~\cite{zhao2025sdrs}  & 82.79 & 82.75 & 70.21 & 46.45 & 0.405 & 0.677 \\
		\textbf{TAME\,\texttt{(N)(B)} (ours)}  & \underline{83.25} & \underline{83.21} & \underline{70.68} & \underline{46.61} & \underline{0.394} & 0.661 \\
		\midrule
		Self-MM$^\ast$\,\texttt{(U)(E)}~\cite{yu2021learning} & 82.61 & 82.54 & 69.31 & 40.13 & 0.398 & 0.651 \\
		ConFEDE$^\ast$\,\texttt{(U)(E)}~\cite{yang2023confede} & 82.73 & 82.68 & 69.36 & 46.21 & 0.394 & 0.653 \\
		SDRS\,\texttt{(N)(E)}~\cite{zhao2025sdrs}           & 84.27 & 84.05 & 71.29 & 46.73 & 0.385 & \underline{0.707} \\
		\textbf{TAME\,\texttt{(N)(E)} (ours)}  & \textbf{85.57} & \textbf{85.50} & \textbf{72.21} & \textbf{48.80} & \textbf{0.367} & \textbf{0.711} \\
		\bottomrule
	\end{tabular}
\end{table}

	We compare TAME against a representative set of multimodal sentiment analysis methods spanning four families: (i)~classical fusion baselines including TFN~\cite{zadeh2017tensor}, LMF~\cite{liu2018efficient}, MulT~\cite{tsai2019multimodal}, and RAVEN~\cite{wang2019words}; (ii)~disentangled and self-supervised methods including MISA~\cite{hazarika2020misa}, FDMER~\cite{yang2022disentangled}, MMIM~\cite{han2021improving}, Self-MM~\cite{yu2021learning}, and SDRS~\cite{zhao2025sdrs}; (iii)~uncertainty- and Mixture-of-Experts-based methods including EUAR~\cite{gao2024enhanced}, U-MOE~\cite{xiao2026uncertainty}, PAMoE~\cite{huang2025pamoe}, MMoLRE~\cite{zhang2025multimodal}, and EMOE~\cite{fang2025emoe}; and (iv)~other recent approaches including PMR~\cite{lv2021progressive}, MICA~\cite{liang2021attention}, IMDer~\cite{wang2023incomplete}, GCNet~\cite{lian2023gcnet}, TCTR~\cite{yang2026tctr}, DLF~\cite{wang2025dlf}, DEVA~\cite{wu2025enriching}, MFN~\cite{zadeh2018memory}, Graph-MFN~\cite{zadeh2018multimodal}, MCTN~\cite{pham2019found}, and DMD~\cite{li2023decoupled}. Baseline results are taken from their published papers under the same training protocol.  Performance comparisons are reported in Tables~\ref{tab:mosi_mosei_combined} and~\ref{tab:chsims}.

	\textbf{CMU-MOSI.} TAME with the DeBERTa-v3 backbone establishes new state-of-the-art on Acc-2 of 88.41\%, F1 of 88.41\%, and Acc-7 of 49.42\% under the aligned setting. It also leads on all four metrics under the unaligned setting. It surpasses MoE-uncertainty baselines, including PAMoE, MMoLRE, and EUAR. The MAE under the aligned setting is 0.650, competitive to PAMoE, while TAME outperforms PAMoE on the three classification metrics by substantial margins: 1.37\% on Acc-2, 1.42\% on F1, and 0.72\% on Acc-7. To verify the effectiveness of TAME for MSA, the one‑sample t‑test is conducted on the results on CMU-MOSI database presented in Table~\ref{tab:mosi_mosei_combined}, comparing our method and all baseline results on each metric. TAME significantly outperforms the baseline mean on all metrics under both aligned and unaligned settings (all $p<0.0001$).  For example, on Acc‑2 under aligned setting, the improvement over the baseline mean of 82.21\% yields $t=8.29$, $p<0.0001$; on MAE under aligned setting, TAME is lower than the baseline mean of 0.837 with $t=–4.87$, $p<0.0001$.
	
	\textbf{CMU-MOSEI.} The largest gains appear on this more challenging benchmark. TAME outperforms all compared baselines on all four metrics under both settings. The improvement on Acc-7 is especially notable. Under the aligned setting, it exceeds  EUAR by $1.11\%$ and MMoLRE by $0.23\%$. Under the unaligned setting, it surpasses EMOE by $2.59\%$. MAE drops to 0.499 / 0.497, marking the first sub-0.50 results on this benchmark among all compared methods. Notably, under the aligned setting, TAME reaches an Acc-7 of 56.01\%, exceeding the baseline mean of 52.41\% by 3.6 percentage points, where $t=8.82$, $df=25$, $p<0.0001$. The MAE values of 0.499 and 0.497 are significantly lower than the baseline means of 0.558 and 0.566 with $p<0.0001$.
	
	\textbf{CH-SIMS.} On the Chinese-language CH-SIMS benchmark, TAME outperforms SDRS on five of the six reported metrics under both Chinese language backbones. The only exception occurs under BERT-base-Chinese, where SDRS marginally leads on Corr, 0.677 vs.\ 0.661. With BERT-base-Chinese, TAME\,\texttt{(N)(B)} still improves Acc-2 by $0.46\%$, F1 by $0.46$, Acc-3 by $0.47\%$, Acc-5 by $0.16\%$. It also reduces MAE by $0.011$. With ERNIE-3.0-base, TAME\,\texttt{(N)(E)} achieves the best results on all six metrics: ACC-2 of 85.57\% F1 of 85.50\%, Acc03 of 72.21\%, Acc-5 of 48.80\%, MAE of 0.367, and Corr of 0.711. By comparison, SDRS\,\texttt{(N)(E)} obtains 84.27\%, 84.05\%, 71.29\%, 46.73\%, 0.385, and 0.707, respectively. These results show that the architectural contributions of TAME, including asymmetric MoE organization, text-anchored cross-modal attention, and uncertainty-guided routing, transfer faithfully across both languages and pre-trained language models. 
	
	Overall, across all three benchmarks, TAME consistently delivers leading or highly competitive performance. On the English datasets CMU‑MOSI and CMU‑MOSEI, it sets new state‑of‑the‑art on multiple metrics and achieves the first sub‑0.50 MAE on CMU‑MOSEI, with all improvements significant. On the Chinese CH‑SIMS benchmark, TAME transfers effectively across both language backbones, outperforming the previous best method on nearly all metrics with significance confirmed in most comparisons. These results collectively demonstrate that the proposed asymmetric expert design and uncertainty‑guided routing form a robust and generalisable framework for MSA.
	
	\subsection{Efficiency Analysis}
	\label{ssec:efficiency}
	
	Beyond accuracy, parameter count and computational cost are critical for practical deployment. We compare TAME against EMOE~\cite{fang2025emoe}, the state-of-the-art Mixture-of-Experts MSA baseline, using identical condition: the same V100 32\,GB hardware, a batch size of 16, and the same preprocessing pipelines. The results are summarized in Table~\ref{tab:efficiency}.
	
	\begin{table}[t!]
		\centering
		\caption{Parameter and computational efficiency of TAME against EMOE on CMU-MOSI (Aligned), CMU-MOSEI (Aligned), and CH-SIMS.}
		\label{tab:efficiency}
		\small
		\setlength{\tabcolsep}{3pt}
		\begin{tabular}{llcc}
			\toprule
			\textbf{Dataset} & \textbf{Method} & \textbf{Params (M)} & \textbf{FLOPs (G)} \\
			\midrule
			\multirow{2}{*}{CMU-MOSI (Aligned)}  & EMOE         & 317.30 & 1.4\\
			& \textbf{TAME} &  195.80 & 1.069 \\
			\midrule
			\multirow{2}{*}{CMU-MOSEI (Aligned)} & EMOE         &  361.15 &  1.531\\
			& \textbf{TAME} & 198.59 &1.286\\
				\midrule
			\multirow{2}{*}{CH-SIMS} & EMOE         & 537.78 & 1.489\\
			& \textbf{TAME} & 114.80 &  0.876\\
			\bottomrule
		\end{tabular}
		\vspace{0pt}
	\end{table}
	
	As shown in Table~\ref{tab:efficiency}, TAME consistently requires substantially fewer parameters and lower FLOPs than EMOE across all three benchmarks. On CMU-MOSEI, TAME uses only $198.59$\,M parameters and 1.286G FLOPs, corresponding to reductions of 45.8\% and 16.0\%, respectively. On CMU-MOSI, the parameter count drops by 38.3\% and FLOPs by 23.6\%. The advantage is even more pronounced on CH-SIMS, where TAME achieves reductions of 78.7\% in parameters and 41.2\% in FLOPs. Notably, on the two English benchmarks these savings are realized despite TAME employing a substantially larger language backbone (DeBERTa-v3-base, approximately 184M parameters) compared to EMOE’s BERT-base (approximately 110M). The structural source of this efficiency is the asymmetric Mixture-of-Modality-Experts organization. EMOE assigns one expert per modality plus an additional fusion gate, causing the parameter footprint to grow roughly linearly with the number of modalities. In contrast, TAME uses only two routed cross-modal experts, A$\to$T and V$\to$T, which share text-anchored cross-attention, while the always-on shared text expert directly reuses the language backbone representation. Moreover, the Prophet Heads and the uncertainty-aware router are deliberately lightweight, comprising only a residual MLP, a single bottleneck, and a 2-class logit. The Pyramid Attention Pooling module reuses the cross-modal feature dimension without expanding it. Consequently, TAME achieves expert-level multimodal fusion while keeping the modality-specific overhead far lower than in conventional MoE designs.
	
	\subsection{Ablation study}
		\label{ssec:ablation}

	We conduct ablation experiments on CMU-MOSEI (Unaligned) to quantify the contribution of each key component in TAME.  Four components are examined: (1) Text-Anchored Cross-Modal Attention (TACMA), (2) Uncertainty Estimation via Prophet Heads (UEPH), (3) Uncertainty-Guided Expert Routing (UGER),  and (4) Pyramid Attention Pooling (PAP), which is replaced by temporal mean pooling when ablated. As well, we also ablate the influence of shared expert defined in Equation~\ref{eq:fusion}. Note that removing UEPH necessarily disables UGER due to their architectural dependency. The results are reported in Table~\ref{tab:mechanism} and Table~\ref{tab:ablation}.

		\begin{table}[th!]
		\centering
		\caption{Ablation study on Text-Anchored Cross-Modal Attention of TAME.}
		\label{tab:mechanism}
		\small
		\setlength{\tabcolsep}{3pt}
		\begin{tabular}{l cccc}
			\toprule
			\textbf{Variant} & Acc-2 (\%)$\uparrow$ & F1 (\%)$\uparrow$ & Acc-7 (\%)$\uparrow$ & MAE$\downarrow$ \\
			\midrule
			T1: TAME & \textbf{88.22} & \textbf{88.15} & \textbf{56.49} & \textbf{0.4966} \\
			\midrule
			T2: +A$\leftrightarrow$V branch            & 86.90 & 86.95 & 54.75 & 0.5144 \\
			T3: Text as Query                    & 87.20 & 87.29 & 56.13 & 0.5056 \\
			T4: T2 + T3         & 86.46 & 86.50 & 54.17 & 0.5190 \\
			\bottomrule
		\end{tabular}
		\vspace{0pt}
	\end{table}
	
\textbf{Effects of Text-Anchored Cross-Modal Attention.} 
We investigate the empirical optimality of TACMA via a two-factor design: (i) whether to include the direct A$\leftrightarrow$V branch, and (ii) the query direction,~\ie, whether queries come from non-text modalities or from text. Table~\ref{tab:mechanism} reports the results for the four configurations, T1 through T4.

First, incorporating the A$\leftrightarrow$V branch consistently impairs performance. Under the non-text-query setting, T1$\to$T2 reduces Acc-2 by 1.32\% and Acc-7 by 1.74\%; under the text-query setting, T3$\to$T4 reduces them by 0.74\% and 1.96\%, respectively. Averaging across both factorial cells yields a main-effect drop of 1.03\% in Acc-2 and 1.85\% in Acc-7. This consistent degradation empirically substantiates the linguistic-bottleneck rationale underlying TACMA.

Second, we examine the reverse configuration by swapping the query–key/value roles relative to Eq.~\ref{eq:ca_attn}. Concretely, for each non‑text modality \(m \in \{a, v\}\), we set the queries to \(\mathbf{H}_t\) while the keys and values are drawn from \(\mathbf{H}_m\), yielding two separate MHA modules—one per modality—whose outputs are subsequently processed by the same FFN as in Eq.~\ref{eq:ca_ffn}. This inversion lets text attend to each non‑text modality independently, creating two distinct cross-modal pathways. We observe that using text as the query (T3) underperforms the full T1 baseline, with drops of 1.02\% in Acc‑2 and 0.37\% in Acc‑7. The interaction between the two factors is small and inconsistent in sign.

Third, T4 combining both modifications drops 1.76\% in Acc-2 and 2.32\% in Acc-7 relative to the full T1 baseline. These degradations closely mirror the cumulative sum of the individual losses from T2 and T3, suggesting that the two design choices contribute in an approximately additive manner. Together, these results confirm that TACMA's default topology is a principled, empirically grounded architectural decision rather than an ad-hoc configuration.
	
	\begin{table}[th!]
		\centering
		\caption{Ablation study of TAME. Each row removes one component from the TAME model. Removing UEPH also disables URGE due to architectural dependency.}
		\label{tab:ablation}
		\small
		\begin{tabular}{lcccc}
			\toprule
			\textbf{Variant} & Acc-2 (\%)$\uparrow$ & F1 (\%)$\uparrow$ & Acc-7 (\%)$\uparrow$ & MAE$\downarrow$ \\
			\midrule
			\textbf{TAME} & \textbf{88.22} & \textbf{88.15} & \textbf{56.49} & \textbf{0.4966} \\
			\midrule
			w/o UEPH & 86.76 & 86.85 & 54.60 & 0.5109 \\
			w/o URGE         & 87.12 & 87.09 & 55.76 & 0.5098 \\
			w/o PAP        & 87.64 & 87.71 & 53.94 & 0.5239 \\
			\bottomrule
		\end{tabular}

	\end{table}
	
%
	
	\textbf{Effects of Uncertainty Estimation via Prophet Heads (UEPH).} 
	We first isolate the contribution of the heteroscedastic uncertainty estimation itself by examining the ``w/o UEPH'' variant. Removing only the uncertainty prior still incurs notable degradations: Acc-2 drops by 1.10\%, F1 by 1.06\%, Acc-7 by 0.73\%, and MAE increases by 0.0132. This indicates that the auxiliary GNLL objective in Eq.~\ref{eq:loss} alone, even without routing feedback, already provides meaningful supervisory signals that improve modality-specific representations and enhance overall robustness.
	
	\textbf{Effects of Uncertainty-Guided Expert Routing (URGE).} 
	To assess the value of feeding uncertainty estimates back into the routing decision, we compare ``w/o UEPH'' against ``w/o URGE''. The additional deterioration, namely a further drop of 0.36\% in Acc-2 and 1.16\% in Acc-7, directly quantifies the benefit of incorporating the prior bias \(\mathbf{b} = [-s_a, -s_v]^\top\) into the router as per Eq.~\ref{eq:router_final}. Crucially, the ``w/o URGE'' variant still underperforms the full model substantially despite retaining UEPH, confirming that uncertainty estimates are most effective when explicitly used to modulate routing weights rather than serving only as an auxiliary training signal. Jointly removing both UEPH and URGE causes the largest overall degradations across all variants, with Acc-2 dropping by 1.46\%, F1 by 1.30\%, Acc-7 by 1.89\%, and MAE increasing by 0.0143. This firmly establishes the complete uncertainty-guided routing pipeline as the core driver of TAME's performance.
	
	\textbf{Effects of Pyramid Attention Pooling (PAP).} Replacing PAP with temporal mean pooling (``w/o PAP'') yields a 0.58\% drop in Acc-2 and 0.44\% drop in F1, yet it causes the largest Acc-7 drop of 2.55\% and the largest MAE increase of 0.0273 among the four component ablations. This pattern suggests that the multi-scale temporal aggregation provided by PAP primarily benefits fine-grained sentiment discrimination and regression accuracy rather than coarse polarity classification, positioning PAP as a supporting representation-quality module rather than a core routing mechanism.

\section{Conclusion}
\label{sec:conclusion}

In this paper, we introduced TAME, a Confidence-Guided Mixture of Modality Experts that re‑formulates multimodal sentiment analysis as uncertainty‑guided refinement of a language prior. Its key novelty lies in an asymmetric Mixture‑of‑Modality‑Experts architecture: the text branch acts as an always‑on shared expert, while two cross‑modal experts (A$\to$T, V$\to$T) are selectively engaged via an uncertainty‑aware router. The router, conditioned on per‑modality confidence estimates from lightweight Prophet Heads, implements an “evaluate‑then‑trust” decision that dynamically admits only reliable nonverbal evidence. Experiments on CMU‑MOSI, CMU‑MOSEI, and CH‑SIMS datasets show TAME achieves new state‑of‑the‑art on Acc‑2, F1, and Acc‑7, while while reducing parameter count by up to 45.8\% on the English benchmarks and by 78.7\% on the Chinese benchmark, alongside FLOPs reductions of up to 41.2\%.. The learned routing weights exhibit a near‑perfect negative correlation with prediction uncertainty, directly validating the router’s ability to favor trustworthy modalities. Future work will explore adaptive shared‑expert selection for non‑text‑centric scenarios, lightweight audio–visual alignment, and extension to additional modalities.

\bibliographystyle{IEEEtran}
\bibliography{refs}

@inproceedings{yu2020chsims,
	title={Ch-sims: A chinese multimodal sentiment analysis dataset with fine-grained annotation of modality},
	author={Yu, Wenmeng and Xu, Hua and Meng, Fanyang and Zhu, Yilin and Ma, Yixiao and Wu, Jiele and Zou, Jiyun and Yang, Kaicheng},
	booktitle={Proc. Annu. Meet. Assoc. Comput. Linguist.},
	pages={3718--3727},
	year={2020}
}

@inproceedings{he2023debertav,
	title={De{BERT}aV3: Improving De{BERT}a using {ELECTRA}-Style Pre-Training with Gradient-Disentangled Embedding Sharing},
	author={Pengcheng He and Jianfeng Gao and Weizhu Chen},
	booktitle={Pro. Int. Conf. Learn. Represent.},
	year={2023}
}

@article{zhao2025sdrs,
	title={SDRS: Sentiment-aware disentangled representation shifting for multimodal sentiment analysis},
	author={Zhao, Sicheng and Yang, Zhenhua and Shi, Henglin and Feng, Xiaocheng and Meng, Lingpengkun and Qin, Bing and Yan, Chenggang and Tao, Jianhua and Ding, Guiguang},
	journal={IEEE Trans. Affect. Comput.},
	volume={16},
	number={3},
	pages={1802--1813},
	year={2025},
	publisher={IEEE}
}

@article{sun2021ernie3,
	title={Ernie 3.0: Large-scale knowledge enhanced pre-training for language understanding and generation},
	author={Sun, Yu and Wang, Shuohuan and Feng, Shikun and Ding, Siyu and Pang, Chao and Shang, Junyuan and Liu, Jiaxiang and Chen, Xuyi and Zhao, Yanbin and Lu, Yuxiang and others},
	journal={arXiv preprint arXiv:2107.02137},
	year={2021}
}

@article{xiao2026uncertainty,
	title={Uncertainty-aware Mixture of Experts for Robust Multimodal Sentiment Analysis},
	author={Xiao, Xinyu and Wang, Xuan and Qi, Shuhan and Li, Huale},
	journal={Pattern Recognit.},
	pages={113417},
	year={2026}
}

@article{huang2025pamoe,
	title={PAMoE-MSA: polarity-aware mixture of experts network for multimodal sentiment analysis},
	author={Huang, Changqin and Lin, Zhenheng and Han, Zhongmei and Huang, Qionghao and Jiang, Fan and Huang, Xiaodi},
	journal={Int. J. Multimed. Inf. Retr.},
	volume={14},
	number={1},
	pages={7},
	year={2025}
}

@inproceedings{yang2023confede,
	title={Confede: Contrastive feature decomposition for multimodal sentiment analysis},
	author={Yang, Jiuding and Yu, Yakun and Niu, Di and Guo, Weidong and Xu, Yu},
	booktitle={Proc. Annu. Meet. Assoc. Comput. Linguist.},
	pages={7617--7630},
	year={2023}
}

@inproceedings{fang2025emoe,
  title={Emoe: Modality-specific enhanced dynamic emotion experts},
  author={Fang, Yiyang and Huang, Wenke and Wan, Guancheng and Su, Kehua and Ye, Mang},
  booktitle={Proc. IEEE/CVF Conf. Comput. Vis. Pattern Recognit.},
  pages={14314--14324},
  year={2025}
}

@inproceedings{williams2018recognizing,
  title={Recognizing emotions in video using multimodal DNN feature fusion},
  author={Williams, Jennifer and Kleinegesse, Steven and Comanescu, Ramona and Radu, Oana},
  booktitle={Proc. of Challenge-HML},
  pages={11--19},
  year={2018}
}

@inproceedings{zadeh2017tensor,
  title={Tensor fusion network for multimodal sentiment analysis},
  author={Zadeh, Amir and Chen, Minghai and Poria, Soujanya and Cambria, Erik and Morency, Louis-Philippe},
  booktitle={Proc. Conf. Empir. Methods Nat. Lang. Process.},
  pages={1103--1114},
  year={2017}
}

@inproceedings{zadeh2018memory,
  title={Memory fusion network for multi-view sequential learning},
  author={Zadeh, Amir and Liang, Paul Pu and Mazumder, Navonil and Poria, Soujanya and Cambria, Erik and Morency, Louis-Philippe},
  booktitle={Proc. AAAI Conf. Artif. Intell.},
  year={2018}
}

@inproceedings{liu2018efficient,
  title={Efficient low-rank multimodal fusion with modality-specific factors},
  author={Liu, Zhun and Shen, Ying and Lakshminarasimhan, Varun Bharadhwaj and Liang, Paul Pu and Zadeh, AmirAli Bagher and Morency, Louis-Philippe},
  booktitle={Proc. Annu. Meet. Assoc. Comput. Linguist.},
  pages={2247--2256},
  year={2018}
}

@inproceedings{tsai2019multimodal,
  title={Multimodal transformer for unaligned multimodal language sequences},
  author={Tsai, Yao-Hung Hubert and Bai, Shaojie and Liang, Paul Pu and Kolter, J Zico and Morency, Louis-Philippe and Salakhutdinov, Ruslan},
  booktitle={Proc. Annu. Meet. Assoc. Comput. Linguist.},
  pages={6558--6569},
  year={2019}
}

@inproceedings{hazarika2020misa,
  title={Misa: Modality-invariant and-specific representations for multimodal sentiment analysis},
  author={Hazarika, Devamanyu and Zimmermann, Roger and Poria, Soujanya},
  booktitle={Proc. ACM Int. Conf. Multimed.},
  pages={1122--1131},
  year={2020}
}

@inproceedings{yang2022disentangled,
  title={Disentangled representation learning for multimodal emotion recognition},
  author={Yang, Dingkang and Huang, Shuai and Kuang, Haopeng and Du, Yangtao and Zhang, Lihua},
  booktitle={Proc. ACM Int. Conf. Multimed.},
  pages={1642--1651},
  year={2022}
}

@inproceedings{lin2017feature,
  title={Feature pyramid networks for object detection},
  author={Lin, Tsung-Yi and Doll{\'a}r, Piotr and Girshick, Ross and He, Kaiming and Hariharan, Bharath and Belongie, Serge},
  booktitle={Proc. IEEE/CVF Conf. Comput. Vis. Pattern Recognit.},
  pages={2117--2125},
  year={2017}
}

@inproceedings{yu2021learning,
  title={Learning modality-specific representations with self-supervised multi-task learning for multimodal sentiment analysis},
  author={Yu, Wenmeng and Xu, Hua and Yuan, Ziqi and Wu, Jiele},
  booktitle={Proc. AAAI Conf. Artif. Intell.},
  pages={10790--10797},
  year={2021}
}

@inproceedings{han2021improving,
  title={Improving multimodal fusion with hierarchical mutual information maximization for multimodal sentiment analysis},
  author={Han, Wei and Chen, Hui and Poria, Soujanya},
  booktitle={Proc. Conf. Empir. Methods Nat. Lang. Process.},
  pages={9180--9192},
  year={2021}
}

@inproceedings{li2023decoupled,
  title={Decoupled multimodal distilling for emotion recognition},
  author={Li, Yong and Wang, Yuanzhi and Cui, Zhen},
  booktitle={Proc. IEEE/CVF Conf. Comput. Vis. Pattern Recognit.},
  pages={6631--6640},
  year={2023}
}

@inproceedings{pham2019found,
  title={Found in translation: Learning robust joint representations by cyclic translations between modalities},
  author={Pham, Hai and Liang, Paul Pu and Manzini, Thomas and Morency, Louis-Philippe and P{\'o}czos, Barnab{\'a}s},
  booktitle={Proc. AAAI Conf. Artif. Intell.},
  pages={6892--6899},
  year={2019}
}

@inproceedings{zadeh2018multimodal,
  title={Multimodal language analysis in the wild: Cmu-mosei dataset and interpretable dynamic fusion graph},
  author={Zadeh, AmirAli Bagher and Liang, Paul Pu and Poria, Soujanya and Cambria, Erik and Morency, Louis-Philippe},
  booktitle={Proc. Annu. Meet. Assoc. Comput. Linguist.},
  pages={2236--2246},
  year={2018}
}

@inproceedings{wu2025enriching,
  title={Enriching multimodal sentiment analysis through textual emotional descriptions of visual-audio content},
  author={Wu, Sheng and He, Dongxiao and Wang, Xiaobao and Wang, Longbiao and Dang, Jianwu},
  booktitle={Proc. AAAI Conf. Artif. Intell.},
  pages={1601--1609},
  year={2025}
}

@inproceedings{wang2025dlf,
  title={DLF: Disentangled-language-focused multimodal sentiment analysis},
  author={Wang, Pan and Zhou, Qiang and Wu, Yawen and Chen, Tianlong and Hu, Jingtong},
  booktitle={Proc. AAAI Conf. Artif. Intell.},
  pages={21180--21188},
  year={2025}
}

@article{lian2023gcnet,
  title={Gcnet: Graph completion network for incomplete multimodal learning in conversation},
  author={Lian, Zheng and Chen, Lan and Sun, Licai and Liu, Bin and Tao, Jianhua},
  journal={IEEE Trans. Pattern Anal. Mach. Intell.},
  volume={45},
  number={7},
  pages={8419--8432},
  year={2023},
  publisher={IEEE}
}

@inproceedings{wang2023incomplete,
  title={Incomplete multimodality-diffused emotion recognition},
  author={Wang, Yuanzhi and Li, Yong and Cui, Zhen},
  booktitle={Proc. Adv. Neural Inf. Process. Syst.},
  volume={36},
  pages={17117--17128},
  year={2023}
}

@inproceedings{liang2021attention,
  title={Attention is not enough: Mitigating the distribution discrepancy in asynchronous multimodal sequence fusion},
  author={Liang, Tao and Lin, Guosheng and Feng, Lei and Zhang, Yan and Lv, Fengmao},
  booktitle={Proc. IEEE Int. Conf. Comput. Vis.},
  pages={8148--8156},
  year={2021}
}

@inproceedings{lv2021progressive,
  title={Progressive modality reinforcement for human multimodal emotion recognition from unaligned multimodal sequences},
  author={Lv, Fengmao and Chen, Xiang and Huang, Yanyong and Duan, Lixin and Lin, Guosheng},
  booktitle={Proc. IEEE/CVF Conf. Comput. Vis. Pattern Recognit.},
  pages={2554--2562},
  year={2021}
}

@inproceedings{wang2019words,
  title={Words can shift: Dynamically adjusting word representations using nonverbal behaviors},
  author={Wang, Yansen and Shen, Ying and Liu, Zhun and Liang, Paul Pu and Zadeh, Amir and Morency, Louis-Philippe},
  booktitle={Proc. AAAI Conf. Artif. Intell.},
  pages={7216--7223},
  year={2019}
}

@article{yang2026tctr,
  title={TCTR: Text-Guided Contrastive Learning with Token-Level Reconstruction Network for missing modalities in multimodal sentiment analysis},
  author={Yang, Zhihao and He, Qing and Yu, Minghao and Du, Nisuo and Lu, Yijie},
  journal={Inf. Fusion.},
  volume={126},
  pages={103571},
  year={2026},
  publisher={Elsevier}
}

@article{zadeh2016mosi,
  title={Mosi: multimodal corpus of sentiment intensity and subjectivity analysis in online opinion videos},
  author={Zadeh, Amir and Zellers, Rowan and Pincus, Eli and Morency, Louis-Philippe},
  journal={arXiv preprint arXiv:1606.06259},
  year={2016}
}

@inproceedings{devlin2019bert,
  title={Bert: Pre-training of deep bidirectional transformers for language understanding},
  author={Devlin, Jacob and Chang, Ming-Wei and Lee, Kenton and Toutanova, Kristina},
  booktitle={Proc. of NAACL HLT},
  pages={4171--4186},
  year={2019}
}

@inproceedings{vaswani2017attention,
  title={Attention is all you need},
  author={Vaswani, Ashish and Shazeer, Noam and Parmar, Niki and Uszkoreit, Jakob and Jones, Llion and Gomez, Aidan N and Kaiser, {\L}ukasz and Polosukhin, Illia},
  booktitle={Proc. Adv. Neural Inf. Process. Syst.},
  volume={30},
  year={2017}
}

@inproceedings{loshchilov2019decoupled,
  title     = {Decoupled Weight Decay Regularization},
  author    = {Loshchilov, Ilya and Hutter, Frank},
  booktitle = {Proc. Int. Conf. Learn. Representations},
   pages={1--10},
  year      = {2019}
}

@article{jacobs1991adaptive,
	title={Adaptive mixtures of local experts},
	author={Jacobs, Robert A and Jordan, Michael I and Nowlan, Steven J and Hinton, Geoffrey E},
	journal={Neural Comput.},
	volume={3},
	number={1},
	pages={79--87},
	year={1991},
	publisher={MIT Press}
}

@inproceedings{dai2024deepseekmoe,
  title={Deepseekmoe: Towards ultimate expert specialization in mixture-of-experts language models},
  author={Dai, Damai and Deng, Chengqi and Zhao, Chenggang and Xu, RX and Gao, Huazuo and Chen, Deli and Li, Jiashi and Zeng, Wangding and Yu, Xingkai and Wu, Yu and others},
  booktitle={Proc. Annu. Meet. Assoc. Comput. Linguist.},
  pages={1280--1297},
  year={2024}
}

@article{jiang2024mixtral,
  title={Mixtral of experts},
  author={Jiang, Albert Q and Sablayrolles, Alexandre and Roux, Antoine and Mensch, Arthur and Savary, Blanche and Bamford, Chris and Chaplot, Devendra Singh and Casas, Diego de las and Hanna, Emma Bou and Bressand, Florian and others},
  journal={arXiv preprint arXiv:2401.04088},
  year={2024}
}

@article{lin2026moe,
  title={Moe-llava: Mixture of experts for large vision-language models},
  author={Lin, Bin and Tang, Zhenyu and Ye, Yang and Huang, Jinfa and Zhang, Junwu and Pang, Yatian and Jin, Peng and Ning, Munan and Luo, Jiebo and Yuan, Li},
  journal={IEEE Trans. Multimedia},
  year={2026},
  publisher={IEEE}
}

@article{han2022trusted,
  title={Trusted multi-view classification with dynamic evidential fusion},
  author={Han, Zongbo and Zhang, Changqing and Fu, Huazhu and Zhou, Joey Tianyi},
  journal={IEEE Trans. Pattern Anal. Mach. Intell.},
  volume={45},
  number={2},
  pages={2551--2566},
  year={2022},
  publisher={IEEE}
}

@inproceedings{hu2018squeeze,
  title={Squeeze-and-excitation networks},
  author={Hu, Jie and Shen, Li and Sun, Gang},
  booktitle={Proc. IEEE/CVF Conf. Comput. Vis. Pattern Recognit.},
  pages={7132--7141},
  year={2018}
}

@article{das2023multimodal,
  title={Multimodal sentiment analysis: a survey of methods, trends, and challenges},
  author={Das, Ringki and Singh, Thoudam Doren},
  journal={ACM Computing Surveys},
  volume={55},
  number={13s},
  pages={1--38},
  year={2023},
  publisher={ACM New York, NY}
}

@incollection{wharton2024non,
  title={Non-verbal communication and context: Multi-modality in interaction},
  author={Wharton, Tim and Madella, Pauline},
  booktitle={The Cambridge Handbook of Language in Context},
  pages={419--436},
  year={2024},
  publisher={Cambridge University Press}
}

@inproceedings{he2016deep,
	  title={Deep Residual Learning for Image Recognition},
	author={He, Kaiming and Zhang, Xiangyu and Ren, Shaoqing and Sun, Jian},
	booktitle={Proc. IEEE/CVF Conf. Comput. Vis. Pattern Recognit.},
	pages={770-778},
	year={2016}
}

@article{xiao2026multimodal,
	title={Uncertainty-aware Mixture of Experts for Robust Multimodal Sentiment Analysis},
	author={Xiao, Xinyu and Wang, Xuan and Qi, Shuhan and Li, Huale},
	journal={Pattern Recognit.},
	pages = {113417},
	year = {2026}
}

@inproceedings{gao2024enhanced,
	author = {Gao, Zixian and Hu, Disen and Jiang, Xun and Lu, Huimin and Shen, Heng Tao and Xu, Xing},
	title = {Enhanced Experts with Uncertainty-Aware Routing for Multimodal Sentiment Analysis},
	year = {2024},
	booktitle = {Proc. ACM Int. Conf. Multimed.},
	pages = {9650–9659}
}

@inproceedings{zhang2025multimodal,
	author={Zhang, Shuo and Zhang, Jinsong and Zhang, Zhejun and Li, Lei},
	booktitle={Proc. IEEE Int. Conf. Multim. Expo.}, 
	title={Multimodal Mixture of Low-Rank Experts for Sentiment Analysis and Emotion Recognition}, 
	year={2025}
}

@article{zhang2023multitask,
	title={A multitask learning model for multimodal sarcasm, sentiment and emotion recognition in conversations},
	author={Zhang, Yazhou and Wang, Jinglin and Liu, Yaochen and Rong, Lu and Zheng, Qian and Song, Dawei and Tiwari, Prayag and Qin, Jing},
	journal={Inf. Fusion.},
	volume={93},
	pages={282--301},
	year={2023},
	publisher={Elsevier}
}

@article{huang2024tmbl,
	title={TMBL: Transformer-based multimodal binding learning model for multimodal sentiment analysis},
	author={Huang, Jiehui and Zhou, Jun and Tang, Zhenchao and Lin, Jiaying and Chen, Calvin Yu-Chian},
	journal={Knowl.-Based Syst.},
	volume={285},
	pages={111346},
	year={2024},
	publisher={Elsevier}
}

@article{hullermeier2021aleatoric,
	title={Aleatoric and epistemic uncertainty in machine learning: An introduction to concepts and methods},
	author={H{\"u}llermeier, Eyke and Waegeman, Willem},
	journal={Machine Learning},
	volume={110},
	number={3},
	pages={457--506},
	year={2021},
	publisher={Springer}
}

@article{huan2025muldef,
	title={Muldef: A model-agnostic debiasing framework for robust multimodal sentiment analysis},
	author={Huan, Ruohong and Zhong, Guowei and Chen, Peng and Liang, Ronghua},
	journal={IEEE Trans. Multimedia},
		volume={27},
	pages={2304--2319},
	year={2025},
	publisher={IEEE}
}

@article{huan2024unimf,
	title={Unimf: A unified multimodal framework for multimodal sentiment analysis in missing modalities and unaligned multimodal sequences},
	author={Huan, Ruohong and Zhong, Guowei and Chen, Peng and Liang, Ronghua},
	journal={IEEE Trans. Multimedia},
	volume={26},
	pages={5753--5768},
	year={2024},
	publisher={IEEE}
}

\end{document}